\pdfoutput=1
\documentclass[twoside,11pt]{article}

\usepackage{jmlr2e}

\usepackage{algorithm}
\usepackage{algorithmic}
\usepackage{paralist,amsmath, amssymb}
\usepackage{microtype}
\usepackage{graphicx}
\usepackage{subfigure}
\usepackage{epstopdf}
\usepackage{booktabs} 
\def \A {\mathcal{A}}

\def \X {\mathcal{X}}
\def \F {\mathcal{F}}
\def \T {\mathcal{T}}
\def \E {\mathbb{E}}

\def \x {\mathbf{x}}

\def \u {\mathbf{u}}
\def \e {\mathbf{e}}
\def \g {\mathbf{g}}
\def \y {\mathbf{y}}

\def \B {\mathcal{B}}
\def \SS {\mathcal{S}}

\DeclareMathOperator*{\argmin}{argmin}

\newtheorem{thm}{Theorem}
\newtheorem{lem}{Lemma}

\newtheorem{assum}{Assumption}

\makeatletter
\newcommand\figcaption{\def\@captype{figure}\caption}
\newcommand\tabcaption{\def\@captype{table}\caption}
\makeatother

\begin{document}

\title{Online Strongly Convex Optimization with Unknown Delays}

\author{\name Yuanyu Wan \email wanyy@lamda.nju.edu.cn\\
\addr National Key Laboratory for Novel Software Technology\\
       Nanjing University, Nanjing 210023, China
\AND
\name Wei-Wei Tu \email tuweiwei@4paradigm.com\\
\addr 4Paradigm Inc., Beijing 100000, China
\AND
        \name Lijun Zhang \email zhanglj@lamda.nju.edu.cn\\
       \addr National Key Laboratory for Novel Software Technology\\
       Nanjing University, Nanjing 210023, China}

\maketitle

\begin{abstract}
We investigate the problem of online convex optimization with unknown delays, in which the feedback of a decision arrives with an arbitrary delay. Previous studies have presented a delayed variant of online gradient descent (OGD), and achieved the regret bound of $O(\sqrt{T+D})$ by only utilizing the convexity condition, where $D$ is the sum of delays over $T$ rounds. In this paper, we further exploit the strong convexity to improve the regret bound. Specifically, we first extend the delayed variant of OGD for strongly convex functions, and establish a better regret bound of $O(d\log T)$, where $d$ is the maximum delay. The essential idea is to let the learning rate decay with the total number of received feedback linearly. Furthermore, we consider the more challenging bandit setting, and obtain similar theoretical guarantees by incorporating the classical multi-point gradient estimator into our extended method. To the best of our knowledge, this is the first work that solves online strongly convex optimization under the general delayed setting.
\end{abstract}


\section{Introduction}
Online convex optimization (OCO) is a prominent paradigm for sequential decision making, which has been successfully applied to many tasks such as portfolio selection \citep{Blum1999,Portfolio_Agarwal,Portfolio_Luo} and online advertisement \citep{McMahan2013,He2014_ADKDD,Juan17_WWW}. At each round $t$, a player selects a decision $\x_t$ from a convex set $\X$. Then, an adversary chooses a convex loss function $f_t(\x):\X\mapsto\mathbb{R}$, and incurs a loss $f_t(\x_t)$ to the player. The performance of the player is measured by the regret
\[R_T=\sum_{t=1}^Tf_t(\x_t)-\min_{\x\in\X}\sum_{t=1}^Tf_t(\x)\]
which is the gap between the cumulative loss of the player and an optimal fixed decision. 

Online gradient descent (OGD) proposed by \citet{Zinkevich2003} is a standard method for minimizing the regret. For convex functions, \citet{Zinkevich2003} showed that OGD attains an $O(\sqrt{T})$ regret bound. If the functions are strongly convex, \citet{Hazan_2007} proved that OGD can achieve a better regret bound of $O(\log T)$. The $O(\sqrt{T})$ and $O(\log T)$ bounds have been proved to be minimax optimal for convex and strongly convex functions, respectively \citep{Abernethy08}. However, the standard OCO assumes that the loss function $f_t(\x)$ is revealed to the player immediately after making the decision $\x_t$, which does not account for the possible delay between the decision and feedback in various practical applications. 
For example, in online advertisement, the decision is about the strategy of serving an ad to a user, and the feedback required to update the decision usually is whether the ad is clicked or not \citep{McMahan2013}. But, after seeing the ad, the user may take some time to give feedback. Moreover, there may not exist a button for the negative feedback, which is not determined unless the user does not click the ad after a sufficiently long period \citep{He2014_ADKDD}. 

To address the above challenge, \citet{Quanrud15} proposed delayed OGD (DOGD) for OCO with unknown delays, and attained the $O(\sqrt{T+D})$ regret bound, where $D$ is the sum of delays over $T$ rounds. Similar to OGD, in each round $t$, DOGD queries the gradient $\nabla f_t(\x_t)$, but according to the delayed setting, it will be received at the end of round $t+d_t-1$ where $d_t\geq1$ is an unknown integer. By the same token, gradients queried in previous rounds may be received at the end of round $t$, and DOGD updates the decision $\x_t$ with the sum of received gradients. Recently, \citet{Li_AISTATS19} further considered the more challenging bandit setting, and proposed delayed bandit gradient descent (DBGD) with $O(\sqrt{T+D})$ regret bound. Specifically, DBGD queries each function $f_t(\x)$ at $n+1$ points where $n$ is the dimensionality, and approximates the gradient by applying the classical $(n+1)$-point gradient estimator \citep{Agarwal2010_COLT} to each received feedback. At the end of round $t$, different from DOGD that only updates the decision $\x_t$ once, DBGD repeatedly updates the decision $\x_t$ with each approximate gradient. While DOGD and DBGD can handle unknown delays for the full information and bandit settings respectively, it remains unclear whether the strong convexity of loss functions can be utilized to achieve a better regret bound.

We notice that \citet{Khashabi16} have tried to exploit the strong convexity for DOGD, but failed because they discovered mistakes in their proof. In this paper, we provide an affirmative answer by proposing a variant of DOGD for strongly convex functions, namely DOGD-SC, which achieves a regret bound of $O(d\log T)$, where $d$ is the maximum delay. To this end, we refine the learning rate used in the original DOGD with a new one that decays with the total number of received feedback linearly, which is able to exploit the strong convexity. For a small $d=O(1)$, our $O(d\log T)$ regret bound is significantly better than the $O(\sqrt{T+D})$ regret bound established by only using the convexity condition. Furthermore, to handle the bandit setting, we propose a bandit variant of DOGD-SC by combining with the $(n+1)$-point gradient estimator \citep{Agarwal2010_COLT}. In each round, we only update the decision once with the sum of approximate gradients, which could be more efficient than DBGD \citep{Li_AISTATS19}. Our theoretical analysis reveals that the bandit variant of DOGD-SC can also obtain the $O(d\log T)$ regret bound for strongly convex functions, which is better than the $O(\sqrt{T+D})$ regret bound of DBGD.
\section{Related Work}
In this section, we briefly review the related work about OCO with delayed feedback, in which the feedback for the decision $\x_t$ is received at the end of round $t+d_t-1$.

\subsection{The Standard OCO}
If $d_t=1$ for all $t\in[T]$, OCO with delayed feedback is reduced to the standard OCO, in which various algorithms have been proposed to minimize the regret under the full information and bandit settings \citep{Online:suvery,Hazan2016}. In the full information setting, by using the gradient of each function, the standard OGD achieves $O(\sqrt{T})$ and $O(\log T)$ regret bounds for convex \citep{Zinkevich2003} and strongly convex functions \citep{Hazan_2007}, respectively. For the bandit setting, where only the function value is available to the player, \citet{Agarwal2010_COLT} proposed to approximate the gradient by querying the function at two points or $n+1$ points. Moreover, they showed that OGD with the approximate gradient can also attain $O(\sqrt{T})$ and $O(\log T)$ regret bounds for convex and strongly convex functions, respectively.

\subsection{OCO with Fixed and Known Delays}
To handle the case that each feedback arrives with a fixed and known delay $d$, i.e., $d_t=d$ for all $t\in[T]$, \citet{Weinberger02_TIT} divide the total $T$ rounds into $d$ subsets $\T_1,\cdots,\T_d$, where $\T_i=\{i,i+d,i+2d,\cdots\}\cap[T]$ for $i=1,\cdots,d$. Over rounds in the subset $\T_i$, they maintain an instance $\A_i$ of a base algorithm $\A$. If the base algorithm $\A$ enjoys a regret bound of $R_\A(T)$ for the standard OCO, \citet{Weinberger02_TIT} showed that their method attains a regret bound of $dR_\A(T/d)$. By setting the base algorithm $\A$ as OGD, the regret bounds could be $O(\sqrt{dT})$ for convex functions and $O(d\log T)$ for strongly convex functions, respectively. However, since this method needs to maintain $d$ instances in total, the space complexity is $d$ times as much as that of the base algorithm. 

By contrast, \citet{Langford09} proposed a more efficient method by simply performing the gradient descent step with a delayed gradient, and also achieved the $O(\sqrt{dT})$ and $O(d\log T)$ regret bounds for convex and strongly convex functions, respectively. Moreover, \citet{Shamir17} combined the fixed delay with the local permutation setting, in which the order of the functions can be modified by a distance of at most $M$. When $M\geq d$, they improved the regret bound to $O(\sqrt{T}(1+\sqrt{d^2/M}))$ for convex functions.

\subsection{OCO with Arbitrary but Time-stamped Delays}
Several previous studies considered another delayed setting, in which each feedback could be delayed by arbitrary rounds, but is time-stamped when it is received. Specifically, \citet{Chris_05} focused on the online classification problem, and analyzed the bound for the number of mistakes. \citet{Joulani13} further proposed to solve OCO under this delayed setting by extending the method of \citet{Weinberger02_TIT}. However, similar to \citet{Weinberger02_TIT}, the method proposed by \citet{Joulani13} needs to maintain multiple instances of a base algorithm, which could be prohibitively resource-intensive. Recently, if each delay $d_t$ grows as $o(t^{c})$ for some known $c<1$, \citet{ICML20_Mertikopoulos} employed the one-point gradient estimator \citep{OBO05} to propose a new method for the bandit setting, and established an expected regret bound of  $\tilde{O}(T^{3/4}+T^{2/3+c/3})$ for convex functions.

\subsection{OCO with Unknown Delays} 
Furthermore, \citet{Quanrud15} considered a more general delayed setting, in which each feedback could be delayed arbitrarily and the time stamp of each feedback could also be unknown, and proposed an efficient method called DOGD. The main idea of DOGD is to query the gradient $\nabla f_t(\x_t)$ at each round $t$, and update the decision $\x_t$ with the sum of those gradients queried at the set of rounds $\F_t=\{k|k+d_k-1=t\}$. Different from \citet{Joulani13}, DOGD enjoys the $O(\sqrt{T+D})$ regret bound without any assumption about delays, where $D$ is the sum of delays over $T$ rounds. \citet{Khashabi16} tried to improve the regret bound of DOGD for strongly convex functions, but did not provide a rigorous analysis. Recently, \citet{Li_AISTATS19} proposed DBGD to handle the more challenging bandit setting. In each round $t$, DBGD queries the function $f_t(\x)$ at $n+1$ points, and repeatedly updates the decision $\x_t$ with each approximate gradient computed by applying the $(n+1)$-point gradient estimator \citep{Agarwal2010_COLT} to each feedback received from the set of rounds $\F_t=\{k|k+d_k-1=t\}$. This method also attains a regret bound of $O(\sqrt{T+D})$, but needs to update the decision $|\F_t|$ times in each round $t$.

If the feedback of each decision $\x_t$ is the entire loss function $f_t(\x)$, \citet{Joulani16} provided an algorithmic framework for extending a base algorithm to the delayed setting. By combining the proposed framework with adaptive online algorithms \citep{McMahan10,Duchi2011}, they improved the $O(\sqrt{T+D})$ regret bound to a data-dependent one. 
If the decision set is unbounded and the order of the received feedback keeps the same as the case without delay, an adaptive algorithm and the data-dependent regret bound for the delayed setting were already presented by \citet{McMahan14}. In the worst case, these data-dependent regret bounds would reduce to $O(\sqrt{T+D})$ or $O(\sqrt{dT})$ where $d$ is the maximum delay, which cannot benefit from the strong convexity.

Although there are many studies about OCO with unknown delays, it remains unclear whether the strong convexity can be utilized to improve the regret bound. This paper provides an affirmative answer by establishing the $O(d\log T)$ regret bound for strongly convex functions.

\section{Main Results}
In this section, we first present DOGD-SC, a variant of DOGD for strongly convex functions, which improves the regret bound. Then, we extend DOGD-SC to the bandit setting.
\subsection{DOGD-SC with Improved Regret}
Following previous studies \citep{Online:suvery,Hazan2016}, we introduce some common assumptions.
\begin{assum}
\label{assum1}
Each loss function $f_t(\x)$ is $L$-Lipschitz over $\X$, i.e., $|f_t(\x)-f_t(\y)|\leq L\|\x-\y\|$, for any $\x,\y\in\X$, where $\|\cdot\|$ denotes the Euclidean norm.
\end{assum}
\begin{assum}
\label{assum2}
The radius of the convex decision set $\X$ is bounded by $R$, i.e., $\|\x\|\leq R$, for any $\x\in\X$.
\end{assum}
\begin{assum}
\label{assum4}
Each loss function $f_t(\x)$ is $\beta$-strongly convex over $\X$, i.e., for any $\x,\y\in\X$ \[f_t(\y)\geq f_t(\x)+\nabla f_t(\x)^\top(\y-\x)+\frac{\beta}{2}\|\x-\y\|^2.\]
\end{assum}
To handle OCO with unknown delays, DOGD \citep{Quanrud15} first arbitrarily chooses $\x_1$ from $\X$. In each round $t$, it queries the gradient $\g_t=\nabla f_t(\x_t)$, and then receives the gradient queried in the set of rounds $\F_t=\{k|k+d_k-1=t\}$. If $|\F_t|=0$, DOGD keeps the decision unchanged as $\x_{t+1}=\x_t$. Otherwise, it updates the decision with the sum of gradients received at this round as
\[\x_{t+1}=\Pi_{\X}\left(\x_{t}-\eta_t\sum_{k\in\F_t}\g_k\right)\]
where $\Pi_{\X}(\y)=\argmin_{\x\in\X}\|\x-\y\|$ for any vector $\y$ is the projection operation. According to \citet{Quanrud15}, DOGD attains a regret bound of $O(\sqrt{T+D})$ by using a constant learning rate $\eta_t=O(1/\sqrt{T+D})$ for all $t\in[T]$, where $D$ is the sum of delays and can be estimated on the fly via the standard ``doubling trick" \citep{P_book_2006}.
\begin{algorithm}[t]
\caption{DOGD-SC}
\label{alg1}
\begin{algorithmic}[1]
\STATE \textbf{Initialization:} Choose an arbitrary vector $\x_1\in\X$ and set $h_0=0$
\FOR{$t=1,2,\cdots,T$}
\STATE Query $\g_t=\nabla f_t(\x_t)$
\STATE $h_t=h_{t-1}+\frac{|\F_t|\beta}{2}$
\STATE $\x_{t+1}=\left\{
\begin{aligned}
&\Pi_{\X}\left(\x_{t}-\frac{1}{h_t}\sum_{k\in\F_t}\g_k\right)\text{ if } |\F_t|>0\\
&\x_{t}\quad\quad\quad\quad\quad\quad\quad\quad~~\text{ otherwise}
\end{aligned}\right.$
\ENDFOR
\end{algorithmic}
\end{algorithm}

However, the constant learning rate cannot utilize the strong convexity of the loss functions. In the standard OCO where $\F_t=\{t\}$ for any $t\in[T]$, \citet{Hazan_2007} have established the $O(\log T)$ regret bound for $\beta$-strongly convex functions by setting $\eta_t=1/(\beta t)$. A significant property of the learning rate is that the inverse of $\eta_t$ is increasing by the modulus of the strong convexity of $f_t(\x)$ per round, i.e.,
\begin{equation}
\label{eta_hazan}
\frac{1}{\eta_{t+1}}-\frac{1}{\eta_{t}}=\beta.
\end{equation}
Inspired by (\ref{eta_hazan}), we initialize $\frac{1}{\eta_{0}}=0$ and update it as
\[\frac{1}{\eta_{t+1}}=\frac{1}{\eta_{t}}+\frac{|\F_t|\beta}{2}\]
where $|\F_t|\beta$ is the modulus of the strong convexity of $\sum_{k\in\F_t}f_{k}(\x)$, and the constant $1/2$ is essential for our analysis. Let $h_t=1/\eta_t$ for $t=0,\cdots,T$. The detailed procedures for strongly convex functions are summarized in Algorithm \ref{alg1}, which is named as DOGD for strongly convex functions (DOGD-SC).

Let $d=\max\{d_t|t=1,\cdots,T\}$ denote the maximum delay. Since there could exist some gradients that arrive after the round $T$, we also define $\F_t=\{k|k+d_k-1=t\}$ for any $t=T+1,\cdots,T+d-1$. Then, we establish the following theorem  regarding the regret of Algorithm \ref{alg1}.
\begin{thm}
\label{thm1}
Under Assumptions \ref{assum1}, \ref{assum2} and \ref{assum4}, Algorithm \ref{alg1} satisfies
\begin{align*}
R_T\leq\left(4\beta RL+5L^2\right)\frac{d}{\beta}\left(1+\ln\frac{T}{|\F_s|}\right)
\end{align*}
where $s=\min\left\{t|t\in[T+d-1],|\F_t|>0\right\}$.
\end{thm}
From Theorem \ref{thm1}, the regret bound of Algorithm \ref{alg1} is on the order of $O(d\log T)$, which is better than the $O(\sqrt{T+D})$ regret bound established by \citet{Quanrud15} as long as $d<\sqrt{T+D}/\log T$. Moreover, if $d=O(1)$, our $O(d\log T)$ regret bound is on the same order as the $O(\log T)$ bound for OCO without delay. We note that \citet{Khashabi16} have tried to use the strong convexity by setting $\eta_t=\frac{2}{\beta t|\F_t|}$. However, in this way, there could exist some rounds such that $(t+1)|\F_{t+1}|\leq t|\F_t|$ and
\[\frac{1}{\eta_{t+1}}-\frac{1}{\eta_{t}}=\frac{\beta}{2}((t+1)|\F_{t+1}|-t|\F_t|)\leq0\]
which makes the proof of their Theorem 3.1 problematic.
\subsection{Algorithm for Bandit Setting}
To handle the bandit setting, following previous studies \citep{Agarwal2010_COLT,Saha_2011}, we further introduce two assumptions, as follows.
\begin{assum}
\label{assumb1}
Let $\B^n$ denote the unit Euclidean ball centered at the origin in $\mathbb{R}^n$. There exists a constant $r$ such that $r\B^n\subseteq\X$.
\end{assum}
\begin{assum}
\label{assumb2}
Each loss function $f_t(\x)$ is $\alpha$-smooth over $\X$, i.e., for any $\x,\y\in\X$
\[f_t(\mathbf{y})\leq f_t(\x)+\nabla f_t(\x)^{\top}(\mathbf{y}-\x)+\frac{\alpha}{2}\|\mathbf{y}-\x\|^2.\]
\end{assum}
In the bandit setting, since only the function value is available to the player instead of the gradient, the problem becomes more challenging. Fortunately, \citet{Agarwal2010_COLT} have proposed to approximate the gradient by querying the function at two points or $n+1$ points. To avoid the cost of querying the function many times, one may prefer to adopt the two-point gradient estimator. However, as discussed by \citet{Li_AISTATS19}, the two-point gradient estimator would fail in the general delayed setting, because it requires the time stamp of each feedback, which could be unknown.

As a result, we will utilize the $(n+1)$-point gradient estimator in the general delayed setting. Define \[\X_\delta=(1-\delta/r)\X=\{(1-\delta/r)\x|\x\in\X\}\] for some $0<\delta<r$. For a function $f(\x):\X\mapsto\mathbb{R}$ and a point $\x\in\X_\delta$, the $(n+1)$-point gradient estimator queries \[f(\x),f(\x+\delta\e_1),\cdots,f(\x+\delta\e_n)\] 
where $\e_i$ denotes the unit vector with the $i$-th entry equal 1, and estimates the gradient $\nabla f(\x)$ by
\begin{equation}
\label{eq_est_grad}
\tilde{\g}=\frac{1}{\delta}\sum_{i=1}^n(f(\x+\delta\e_i)-f(\x))\e_i.
\end{equation}
Previous studies have proved that the approximate gradient enjoys the following properties.
\begin{lem}
\label{lemb111} (Lemma 4 in \citet{Li_AISTATS19})
If $f(\x):\X\mapsto\mathbb{R}$ is $L$-Lipschitz and $\alpha$-smooth, for any $\x\in\X_\delta$, it holds that
\[\|\tilde{\g}\|\leq \sqrt{n}L \text{ and }\|\tilde{\g}-\nabla f(\x)\|\leq\frac{\sqrt{n}\alpha\delta}{2}\]
where $\tilde{\g}$ is computed as (\ref{eq_est_grad}).
\end{lem}
From Lemma \ref{lemb111}, the $(n+1)$-point gradient estimator can closely approximate the gradient with a small $\delta$.
\begin{algorithm}[t]
\caption{BDOGD-SC}
\label{alg2}
\begin{algorithmic}[1]
\STATE \textbf{Input:} A parameter $\delta> 0$
\STATE \textbf{Initialization:} Choose an arbitrary vector $\x_1\in\X_\delta$ and set $h_0=0$
\FOR{$t=1,2,\cdots,T$}
\STATE Query $f_t(\x_{t}),f_t(\x_{t}+\delta\e_1),\cdots,f_t(\x_{t}+\delta\e_d)$
\STATE $h_t=h_{t-1}+\frac{|\F_t|\beta}{2}$
\STATE $\x_{t+1}=\left\{
\begin{aligned}
&\Pi_{\X_\delta}\left(\x_{t}-\frac{1}{h_t}\sum_{k\in\F_t}\tilde{\g}_k\right)\text{ if } |\F_t|>0\\
&\x_{t}\quad\quad\quad\quad\quad\quad\quad\quad\quad\text{ otherwise}
\end{aligned}\right.$\\
where $\tilde{\g}_k=\frac{1}{\delta}\sum_{i=1}^n(f_k(\x_{k}+\delta\e_i)-f_k(\x_k))\e_i$
\ENDFOR
\end{algorithmic}
\end{algorithm}

To combine Algorithm \ref{alg1} with the $(n+1)$-point gradient estimator, we need to make three changes as follows. First, at each round $t$, the player queries the function $f_t(\x)$ at $n+1$ points $\x_{t},\x_{t}+\delta\e_1,\cdots,\x_{t}+\delta\e_{n}$, instead of querying the gradient $\nabla f_t(\x_t)$. In this way, the feedback arrives at the end of round $t$ is \[\left\{\{f_k(\x_{k}+\delta\e_i)\}_{i=0}^n|k+d_k-1=t\right\}\] where $\e_0$ is defined as the zero vector.
According to (\ref{eq_est_grad}), we can approximate the gradient $\nabla f_k(\x_k)$ as
\[\tilde{\g}_k=\frac{1}{\delta}\sum_{i=1}^n(f_k(\x_{k}+\delta\e_i)-f_k(\x_k))\e_i\]
for $k\in\F_t$.
Therefore, the second change is to update $\x_t$ with the sum of gradients estimated from the feedback. Moreover, to ensure that $\x_{t}+\delta\e_1,\cdots,\x_{t}+\delta\e_{n}\in\X$, the third change is to limit $\x_t$ in the set $\X_\delta$ for all $t\in[T]$. Combining the second and third changes, we update the decision as
\[\x_{t+1}=\left\{
\begin{aligned}
&\Pi_{\X_\delta}\left(\x_{t}-\frac{1}{h_t}\sum_{k\in\F_t}\tilde{\g}_k\right)\text{ if } |\F_t|>0,\\
&\x_{t}\quad\quad\quad\quad\quad\quad\quad\quad\quad\text{ otherwise}.
\end{aligned}\right.\]
Note that computing $\tilde{\g}_k$ and $\sum_{k\in\F_t}\tilde{\g}_k$ does not require the time stamp of each feedback. 
The detailed procedures for the bandit setting are summarized in Algorithm \ref{alg2}, which is named as a bandit variant of DOGD-SC (BDOGD-SC).

Since there are $n+1$ decisions selected in each round, following \citet{Agarwal2010_COLT}, the regret is redefined as the average regret
\[\tilde{R}_T=\frac{1}{n+1}\sum_{t=1}^T\sum_{i=0}^{n}f_t(\x_{t}+\delta\e_i)-\min_{\x\in\X}\sum_{t=1}^Tf_t(\x).\]We establish the following theorem regarding the average regret of Algorithm \ref{alg2}.
\begin{thm}
\label{thm2}
Let $\tilde{L}=L+\frac{\sqrt{n}\alpha\delta}{2}$ and $\delta=\frac{c\ln T}{T}$, where $c>0$ is a constant such that $\delta<r$. Under Assumptions \ref{assum1}, \ref{assum2}, \ref{assum4}, \ref{assumb1} and \ref{assumb2}, Algorithm \ref{alg2} ensures
\begin{align*}
\tilde{R}_T\leq&\left(4\beta R\tilde{L}+5\tilde{L}^2\right)\frac{d}{\beta}\left(1+\ln\frac{T}{|\F_s|}\right)+\sqrt{n}c\alpha R\ln T\\
&+\frac{cLR\ln T}{r}+cL\ln T
\end{align*}
where $s=\min\left\{t|t\in[T+d-1],|\F_t|>0\right\}$.
\end{thm}
According to Theorem \ref{thm2}, the regret bound of our Algorithm \ref{alg2} is also on the order of $O(d\log T)$, which is better than the $O(\sqrt{T+D})$ regret bound of DBGD \citep{Li_AISTATS19} as long as $d<\sqrt{T+D}/\log T$. Furthermore, in each round $t$, DBGD updates $|F_t|$ times to obtain $\x_{t+1}$, which is more expensive than our Algorithm \ref{alg2}.

Besides Algorithm \ref{alg2}, an algorithm based on the two-point gradient estimator is developed in the appendix, which can handle the case where the time stamp of each feedback is known.

\section{Analysis}
In this section, we only provide the proof of Theorem \ref{thm1}, and the omitted proofs can be found in the appendix.
\subsection{Preliminaries}
According to Algorithm \ref{alg1}, there could exist some feedback that arrives after the round $T$ and is not used to update the decision. However, it is useful for the analysis. Therefore, we perform a virtual update as
\begin{align*}
&h_t=h_{t-1}+\frac{|\F_t|\beta}{2},\\
&\x_{t+1}=\left\{
\begin{aligned}
&\Pi_{\X}\left(\x_{t}-\frac{1}{h_t}\sum_{k\in\F_t}\g_k\right)\text{ if } |\F_t|>0\\
&\x_{t}\quad\quad\quad\quad\quad\quad\quad\quad~~\text{ otherwise}
\end{aligned}\right.
\end{align*}
for $t\in[T+1,T+d-1]$. 

Then, for any $t\in[T+d-1]$, we define 
\begin{equation}
\label{eq_pre_1}
\x^\prime_{t+1}=\left\{
\begin{aligned}
&\x_{t}-\frac{1}{h_t}\sum_{k\in\F_t}\g_k\text{ if } |\F_t|>0,\\
&\x_{t}\quad\quad\quad\quad\quad~\text{ otherwise.}
\end{aligned}\right.
\end{equation} 
Moreover, we define $t^\prime=t+d_t-1$ for any $t\in[T]$ and $s=\min\left\{t|t\in[T+d-1],|\F_t|>0\right\}$.
\subsection{Proof of Theorem \ref{thm1}}
Let $\x^\ast=\argmin_{\x\in\X}\sum_{t=1}^Tf_t(\x)$. We have
\begin{equation}
\label{thm1_eq1}
\begin{split}
R_T=&\sum_{t=1}^Tf_t(\x_t) - \sum_{t=1}^Tf_t(\x^\ast)\\
\leq& \sum_{t=1}^T\left(\nabla f_t(\x_t)^\top(\x_t-\x^\ast)-\frac{\beta}{2}\|\x_t-\x^\ast\|^2\right)\\
=&\sum_{t=1}^T\left(\nabla f_t(\x_t)^\top(\x_{t^\prime}-\x^\ast)-\frac{\beta}{2}\|\x_t-\x^\ast\|^2\right)+\sum_{t=1}^T\nabla f_t(\x_t)^\top(\x_{t}-\x_{t^\prime})\\
\leq&\sum_{t=1}^T\left(\nabla f_t(\x_t)^\top(\x_{t^\prime}-\x^\ast)-\frac{\beta}{2}\|\x_t-\x^\ast\|^2\right)+\sum_{t=1}^T L\|\x_{t}-\x_{t^\prime}\|
\end{split}
\end{equation}
where the first inequality is due to Assumption \ref{assum4}, and the last inequality is due to \begin{align*}
\nabla f_t(\x_t)^\top(\x_{t}-\x_{t^\prime})\leq&\|\nabla f_t(\x_t)\|\|\x_{t}-\x_{t^\prime}\|\\
\leq&L\|\x_{t}-\x_{t^\prime}\|.
\end{align*}
To upper bound the right side of (\ref{thm1_eq1}), we introduce the following lemma.
\begin{lem}
\label{lemA}
For any $\x\in\X$, Algorithm \ref{alg1} ensures
\begin{equation}
\label{thm1_eq4}
\begin{split}
\sum_{t=1}^T\left(\nabla f_t(\x_t)^\top(\x_{t^\prime}-\x)-\frac{\beta}{2}\|\x_t-\x\|^2\right)\leq\sum_{t=1}^T\beta R\|\x_t-\x_{t^\prime}\|+\sum_{t=s}^{T+d-1}\frac{d|\F_t|L^2}{2h_t}.
\end{split}
\end{equation}
\end{lem}
Substituting (\ref{thm1_eq4}) into (\ref{thm1_eq1}), we have
\begin{equation}
\label{thm1_eq6-1}
\begin{split}
R_T\leq&\left(\beta R+L\right)\sum_{t=1}^T\|\x_t-\x_{t^\prime}\|+\sum_{t=s}^{T+d-1}\frac{d|\F_t|L^2}{2h_t}.
\end{split}
\end{equation}
According to the definition of $\x_{t+1}^\prime$, for any $t\in[T+d-1]$, it holds that \begin{equation}
\label{eq_temp1}
\sum_{k\in \F_t}\nabla f_k(\x_k)=h_t(\x_t-\x_{t+1}^\prime).
\end{equation}
Moreover, since $\x_{t+1}=\Pi_{\X}(\x_{t+1}^\prime)$, for any $\x\in\X$, we have 
\begin{equation}
\label{eq_temp2}
\|\x_{t+1}-\x\|\leq\|\x_{t+1}^\prime-\x\|.
\end{equation}
Then, it is not hard to verify that
\begin{equation}
\label{eq_lem1}
\begin{split}
\|\x_{t^\prime}-\x_{t}\|\leq&\sum_{i=t}^{t^\prime-1}\|\x_{i+1}-\x_{i}\|\\
\leq&\sum_{i=t}^{t^\prime-1}\|\x_{i+1}^\prime-\x_{i}\|\\
=&\sum_{i=\max(t,s)}^{t^\prime-1}\frac{\|\sum_{k\in\F_i}\nabla f_k(\x_k)\|}{h_i}\\
\leq&\sum_{i=\max(t,s)}^{t^\prime-1}\frac{|\F_i|L}{h_i}
\end{split}
\end{equation}
where the second inequality is due to (\ref{eq_temp2}), the equality is due to (\ref{eq_temp1}), and the last inequality is due to \begin{equation}
\label{eq_temp3}
\left\|\sum_{k\in \F_i}\nabla f_k(\x_k)\right\|\leq\sum_{k\in \F_i}\|\nabla f_k(\x_k)\|\leq|\F_i|L.
\end{equation}

Substituting (\ref{eq_lem1}) into (\ref{thm1_eq6-1}), we have
\begin{equation*}
\begin{split}
R_T\leq&\left(\beta RL+L^2\right)\sum_{t=1}^T\sum_{i=\max(t,s)}^{t^\prime-1}\frac{|\F_i|}{h_i}+\sum_{t=s}^{T+d-1}\frac{d|\F_t|L^2}{2h_t}.
\end{split}
\end{equation*}
Furthermore, we introduce the following lemma.
\begin{lem}
\label{lem2}
Algorithm \ref{alg1} ensures
\[\sum_{t=1}^T\sum_{i=\max(t,s)}^{t^\prime-1}\frac{|\F_i|}{h_i}\leq2d\sum_{t=s}^{T+d-1}\frac{|\F_t|}{h_t}\] and
\[\sum_{t=s}^{T+d-1}\frac{|\F_t|}{2h_t}\leq\frac{1}{\beta}\left(1+\ln\frac{T}{|\F_s|}\right).\]
\end{lem}
Applying Lemma \ref{lem2}, we have
\begin{equation*}
\begin{split}
R_T\leq&\left(\beta RL+L^2\right)2d\sum_{t=s}^{T+d-1}\frac{|\F_t|}{h_t}+\sum_{t=s}^{T+d-1}\frac{d|\F_t|L^2}{2h_t}\\
\leq&\left(4\beta RL+5L^2\right)\frac{d}{\beta}\left(1+\ln\frac{T}{|\F_s|}\right).
\end{split}
\end{equation*}

\subsection{Proof of Lemma \ref{lemA}}
First, we note that
\begin{equation}
\label{thm1_eq2}
\begin{split}
&\sum_{t=1}^T\nabla f_t(\x_t)^\top(\x_{t^\prime}-\x)\\
=&\sum_{t=1}^{T+d-1}\sum_{k\in \F_t}\nabla f_k(\x_k)^\top(\x_{k+d_k-1}-\x)\\
=&\sum_{t=s}^{T+d-1}\sum_{k\in \F_t}\nabla f_k(\x_k)^\top(\x_{k+d_k-1}-\x)\\
=&\sum_{t=s}^{T+d-1}\sum_{k\in \F_t}\nabla f_k(\x_k)^\top(\x_{t}-\x).
\end{split}
\end{equation}
where the last equality is due to $k+d_k-1=t$ for any $k\in\F_t$.

Substituting (\ref{eq_temp1}) into (\ref{thm1_eq2}), we have
\begin{equation}
\label{eq_temp_lemA}
\begin{split}
&\sum_{t=1}^T\nabla f_t(\x_t)^\top(\x_{t^\prime}-\x)\\
=&\sum_{t=s}^{T+d-1}h_t(\x_t-\x_{t+1}^\prime)^\top(\x_{t}-\x)\\
=&\sum_{t=s}^{T+d-1}\frac{h_t}{2}\left(\|\x_t-\x\|^2-\|\x_{t+1}^\prime-\x\|^2+\|\x_t-\x_{t+1}^\prime\|^2\right)\\
=&\sum_{t=s}^{T+d-1}\frac{h_t}{2}\left(\|\x_t-\x\|^2-\|\x_{t+1}^\prime-\x\|^2\right)+\sum_{t=s}^{T+d-1}\frac{\|\sum_{k\in \F_t}\nabla f_k(\x_k)\|^2}{2h_t}\\
\leq&\sum_{t=s}^{T+d-1}\left(\frac{h_t}{2}\left(\|\x_t-\x\|^2-\|\x_{t+1}^\prime-\x\|^2\right)+\frac{|\F_t|^2L^2}{2h_t}\right)\\
\leq&\sum_{t=s}^{T+d-1}\left(\frac{h_t}{2}\left(\|\x_t-\x\|^2-\|\x_{t+1}-\x\|^2\right)+\frac{|\F_t|^2L^2}{2h_t}\right)\\
=&\sum_{t=s+1}^{T+d-1}\|\x_t-\x\|^2\left(\frac{h_{t}}{2}-\frac{h_{t-1}}{2}\right)+\sum_{t=s}^{T+d-1}\frac{|\F_t|^2L^2}{2h_t}+\frac{h_s}{2}\|\x_s-\x\|^2-\frac{h_{T+d-1}\|\x_{T+d}-\x\|^2}{2}
\end{split}
\end{equation}
where the first inequality is due to (\ref{eq_temp3}), and the last inequality is due to (\ref{eq_temp2}).

According to Algorithm \ref{alg1} and the virtual update, it is easy to verify that \[h_t=\frac{\sum_{i=s}^t|\F_i|\beta}{2}\] 
for any $t\in[s,T+d-1]$.

Combining the above equality with (\ref{eq_temp_lemA}), we have
\begin{equation*}
\begin{split}
&\sum_{t=1}^T\nabla f_t(\x_t)^\top(\x_{t^\prime}-\x)\\
\leq&\sum_{t=s+1}^{T+d-1}\|\x_t-\x\|^2\left(\frac{h_{t}}{2}-\frac{h_{t-1}}{2}\right)+\sum_{t=s}^{T+d-1}\frac{|\F_t|^2L^2}{2h_t}+\frac{h_s}{2}\|\x_s-\x\|^2\\
=&\sum_{t=s+1}^{T+d-1}\frac{|\F_t|\beta}{4}\|\x_t-\x\|^2+\sum_{t=s}^{T+d-1}\frac{|\F_t|^2L^2}{2h_t}+\frac{|\F_s|\beta}{4}\|\x_s-\x\|^2\\
=&\sum_{t=s}^{T+d-1}\frac{|\F_t|\beta}{4}\|\x_t-\x\|^2+\sum_{t=s}^{T+d-1}\frac{|\F_t|^2L^2}{2h_t}.
\end{split}
\end{equation*}
Moreover, we have
\begin{equation*}
\begin{split}
&\sum_{t=1}^T\left(\nabla f_t(\x_t)^\top(\x_{t^\prime}-\x)-\frac{\beta}{2}\|\x_t-\x\|^2\right)\\
\leq&\sum_{t=s}^{T+d-1}\frac{|\F_t|\beta}{4}\|\x_t-\x\|^2+\sum_{t=s}^{T+d-1}\frac{|\F_t|^2L^2}{2h_t}-\sum_{t=1}^T\frac{\beta}{2}\|\x_t-\x\|^2\\
=&\sum_{t=s}^{T+d-1}\frac{|\F_t|\beta}{4}\|\x_t-\x\|^2+\sum_{t=s}^{T+d-1}\frac{|\F_t|^2L^2}{2h_t}-\sum_{t=s}^{T+d-1}\sum_{k\in\F_t}\frac{\beta}{2}\|\x_k-\x\|^2\\
\leq&\sum_{t=s}^{T+d-1}\frac{|\F_t|\beta}{4}\|\x_t-\x\|^2+\sum_{t=s}^{T+d-1}\frac{|\F_t|^2L^2}{2h_t}+\sum_{t=s}^{T+d-1}\sum_{k\in\F_t}\frac{\beta}{2}\left(-\frac{\|\x_t-\x\|^2}{2}+\|\x_t-\x_k\|^2\right)\\
=&\sum_{t=s}^{T+d-1}\sum_{k\in\F_t}\frac{\beta}{2}\|\x_t-\x_k\|^2+\sum_{t=s}^{T+d-1}\frac{|\F_t|^2L^2}{2h_t}\\
\leq&\sum_{t=s}^{T+d-1}\sum_{k\in\F_t} \beta R\|\x_t-\x_k\|+\sum_{t=s}^{T+d-1}\frac{|\F_t|^2L^2}{2h_t}
\end{split}
\end{equation*}
where the second inequality is due to \[\|\x_t-\x\|^2\leq 2\|\x_t-\x_k\|^2+2\|\x_k-\x\|^2\]
and the last inequality is due to Assumption \ref{assum2}.

Since $1\leq d_k\leq d$, for any $t\in[T+d-1]$ and $k\in\F_t$, it is easy to verify that \[t-d+1\leq k=t-d_k+1\leq t\]
which implies that $|\F_t|\leq t-(t-d+1)+1=d$ and
\begin{equation*}
\begin{split}
&\sum_{t=1}^T\left(\nabla f_t(\x_t)^\top(\x_{t^\prime}-\x)-\frac{\beta}{2}\|\x_t-\x\|^2\right)\\
\leq&\sum_{t=s}^{T+d-1}\sum_{k\in\F_t} \beta R\|\x_t-\x_k\|+\sum_{t=s}^{T+d-1}\frac{d|\F_t|L^2}{2h_t}\\
=&\sum_{t=s}^{T+d-1}\sum_{k\in\F_t} \beta R\|\x_{k+d_k-1}-\x_k\|+\sum_{t=s}^{T+d-1}\frac{d|\F_t|L^2}{2h_t}\\
=&\sum_{t=1}^T\beta R\|\x_t-\x_{t^\prime}\|+\sum_{t=s}^{T+d-1}\frac{d|\F_t|L^2}{2h_t}
\end{split}
\end{equation*}
where the first equality is due to $k+d_k-1=t$ for any $k\in\F_t$.

\subsection{Proof of Lemma \ref{lem2}}
First, it is not hard to verify that
\begin{align*}
\sum_{t=1}^T\sum_{i=\max(t,s)}^{t^\prime-1}\frac{|\F_i|}{h_i}=&\sum_{t=1}^{s-1}\sum_{i=s}^{t^\prime-1}\frac{|\F_i|}{h_i}+\sum_{t=s}^T\sum_{i=t}^{t^\prime-1}\frac{|\F_i|}{h_i}\\
\leq&\sum_{t=1}^{s-1}\sum_{i=s}^{t^\prime}\frac{|\F_i|}{h_i}+\sum_{t=s}^T\sum_{i=t}^{t^\prime}\frac{|\F_i|}{h_i}\\
\leq&\sum_{t=1}^{s-1}\sum_{i=s}^{T+d-1}\frac{|\F_i|}{h_i}+\sum_{t=s}^T\sum_{i=t}^{t^\prime}\frac{|\F_i|}{h_i}\\
\leq&\sum_{t=1}^{s-1}\sum_{i=s}^{T+d-1}\frac{|\F_i|}{h_i}+\sum_{t=s}^T\sum_{i=t}^{t+d-1}\frac{|\F_i|}{h_i}\\
=&\sum_{t=1}^{s-1}\sum_{i=s}^{T+d-1}\frac{|\F_i|}{h_i}+\sum_{i=0}^{d-1}\sum_{t=s+i}^{T+i}\frac{|\F_t|}{h_t}\\
\leq&\sum_{t=1}^{s-1}\sum_{i=s}^{T+d-1}\frac{|\F_i|}{h_i}+\sum_{i=0}^{d-1}\sum_{t=s}^{T+d-1}\frac{|\F_t|}{h_t}\\
=&(s-1+d)\sum_{t=s}^{T+d-1}\frac{|\F_t|}{h_t}
\end{align*}
where the second inequality is due to $t^\prime=t+d_t-1\leq T+d-1$, and the third inequality is due to $t^\prime=t+d_t-1\leq t+d-1$.

Since $1+d_1-1=d_1\leq d$, we note that $s\leq d$. Therefore, we have $s-1+d\leq2d$ and
\begin{align*}
\sum_{t=1}^T\sum_{i=\max(t,s)}^{t^\prime-1}\frac{|\F_i|}{h_i}\leq2d\sum_{t=s}^{T+d-1}\frac{|\F_t|}{h_t}.
\end{align*}
Then, we continue to prove the second inequality in Lemma \ref{lem2} with the following lemma.
\begin{lem}
\label{lem3}
Let $a_1>0$ and $a_2,\cdots,a_m\geq0$ be real numbers and let $f:(0,+\infty)\mapsto[0,+\infty)$ be a nonincreasing function. Then
\begin{align*}
\sum_{i=1}^ma_if(a_1+\cdots+a_{i})\leq a_1f(a_1)+\int_{a_1}^{a_1+\cdots+a_m}f(x)dx.
\end{align*}
\end{lem}
Let $f(x)=\frac{1}{x}$ and $a_i=|\F_{s+i-1}|$ for any $i\in[T+d-s]$. Then, we have \[a_1+\cdots+a_{T+d-s}=\sum_{t=s}^{T+d-1}|\F_{t}|=T.\]
Because of $h_t=\frac{\sum_{i=s}^t|\F_i|\beta}{2}$ for any $t\in[s,T+d-1]$, we have
\begin{equation}
\label{lem2_eq1}
\begin{split}
\sum_{t=s}^{T+d-1}\frac{|\F_t|}{2h_t}=&\frac{1}{\beta}\sum_{i=1}^{T+d-s}a_if(a_1+\cdots+a_{i})\\
\leq&\frac{1}{\beta}\left(1+\int_{|\F_s|}^{T}\frac{1}{x}dx\right)\\
=&\frac{1}{\beta}\left(1+\ln\frac{T}{|\F_s|}\right)
\end{split}
\end{equation}
where the first inequality is due to Lemma \ref{lem3}.

\subsection{Proof of Lemma \ref{lem3}}
Lemma \ref{lem3} is inspired by Lemma 14 in \citet{Gaillard14}, which provides the following bound
\[\sum_{i=2}^ma_if(a_1+\cdots+a_{i-1})\leq f(a_1)+\int_{a_1}^{a_1+\cdots+a_m}f(x)dx\]
for $a_2,\cdots,a_m\in[0,1]$. It is not hard to prove Lemma \ref{lem3} by slightly modifying the proof of Lemma 14 in \citet{Gaillard14} to deal with $\sum_{i=1}^ma_if(a_1+\cdots+a_{i})$, instead of $\sum_{i=2}^ma_if(a_1+\cdots+a_{i-1})$. We include the proof for completeness.

Let $s_i=a_1+\cdots+a_i$ for any $i\in[m]$. Then, for any $i=2,\cdots,m$, we have
\[a_if(s_i)=\int_{s_{i-1}}^{s_i}f(s_i)dx\leq\int_{s_{i-1}}^{s_i}f(x)dx\]
where the inequality is due to the fact that $f(x)$ is a nonincreasing function.

Then, we have
\begin{align*}
\sum_{i=1}^ma_if(s_i)&=a_1f(a_1)+\sum_{i=2}^ma_if(s_i)\\
&\leq a_1f(a_1)+\int_{s_{1}}^{s_m}f(x)dx.
\end{align*}

\section{Experiments}
In this section, we provide numerical experiments to verify the performance of our DOGD-SC and BDOGD-SC for strongly convex functions.
\begin{figure*}[t]
\centering
\subfigure[Low Delayed Setting]{\includegraphics[width=0.49\textwidth]{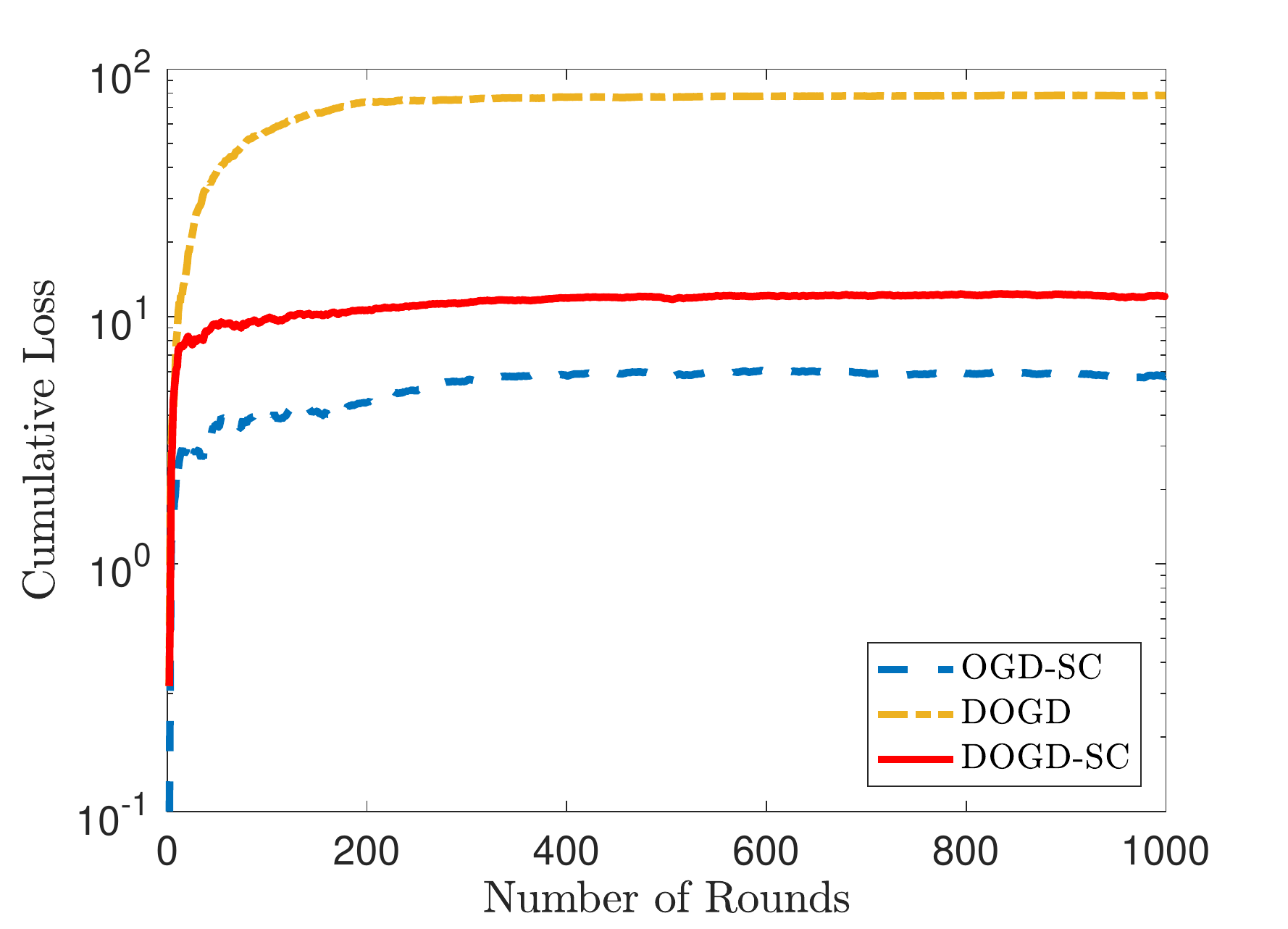}}
\centering
\subfigure[High Delayed Setting]{\includegraphics[width=0.49\textwidth]{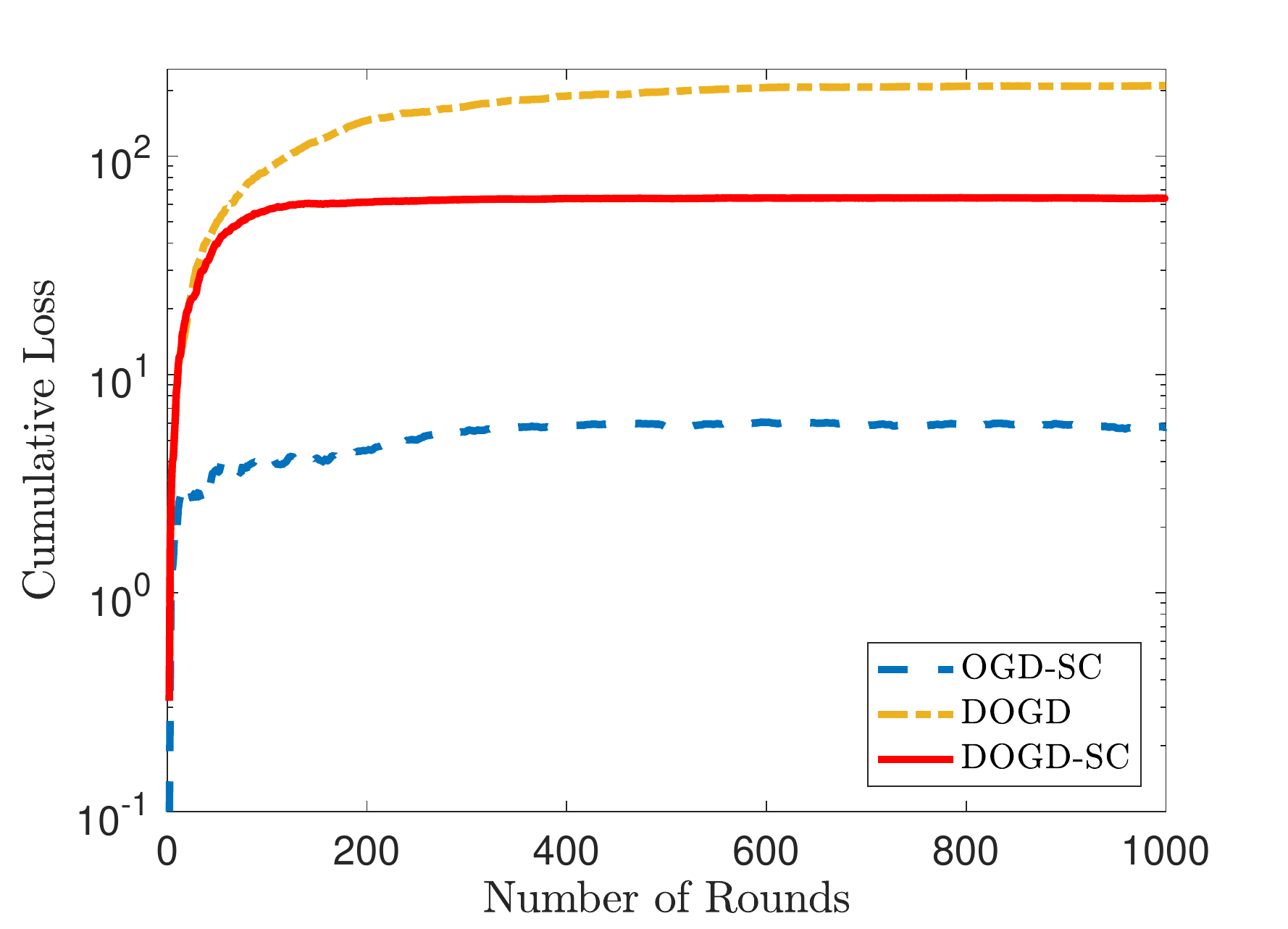}}
\caption{Comparisons of our Algorithm 1 against OGD-SC and DOGD.}
\label{fig1}
\centering
\subfigure[Low Delayed Setting]{\includegraphics[width=0.49\textwidth]{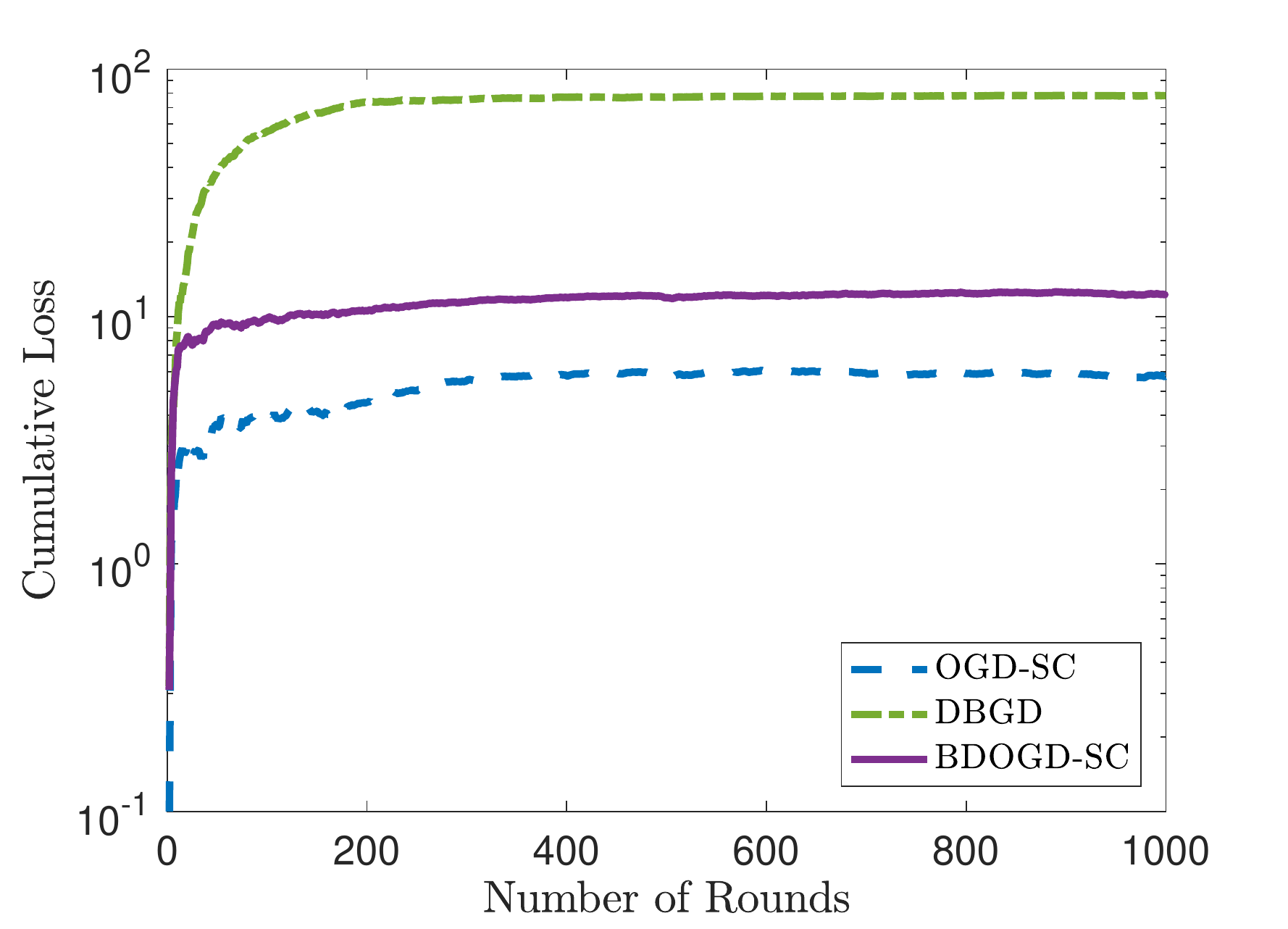}}
\centering
\subfigure[High Delayed Setting]{\includegraphics[width=0.49\textwidth]{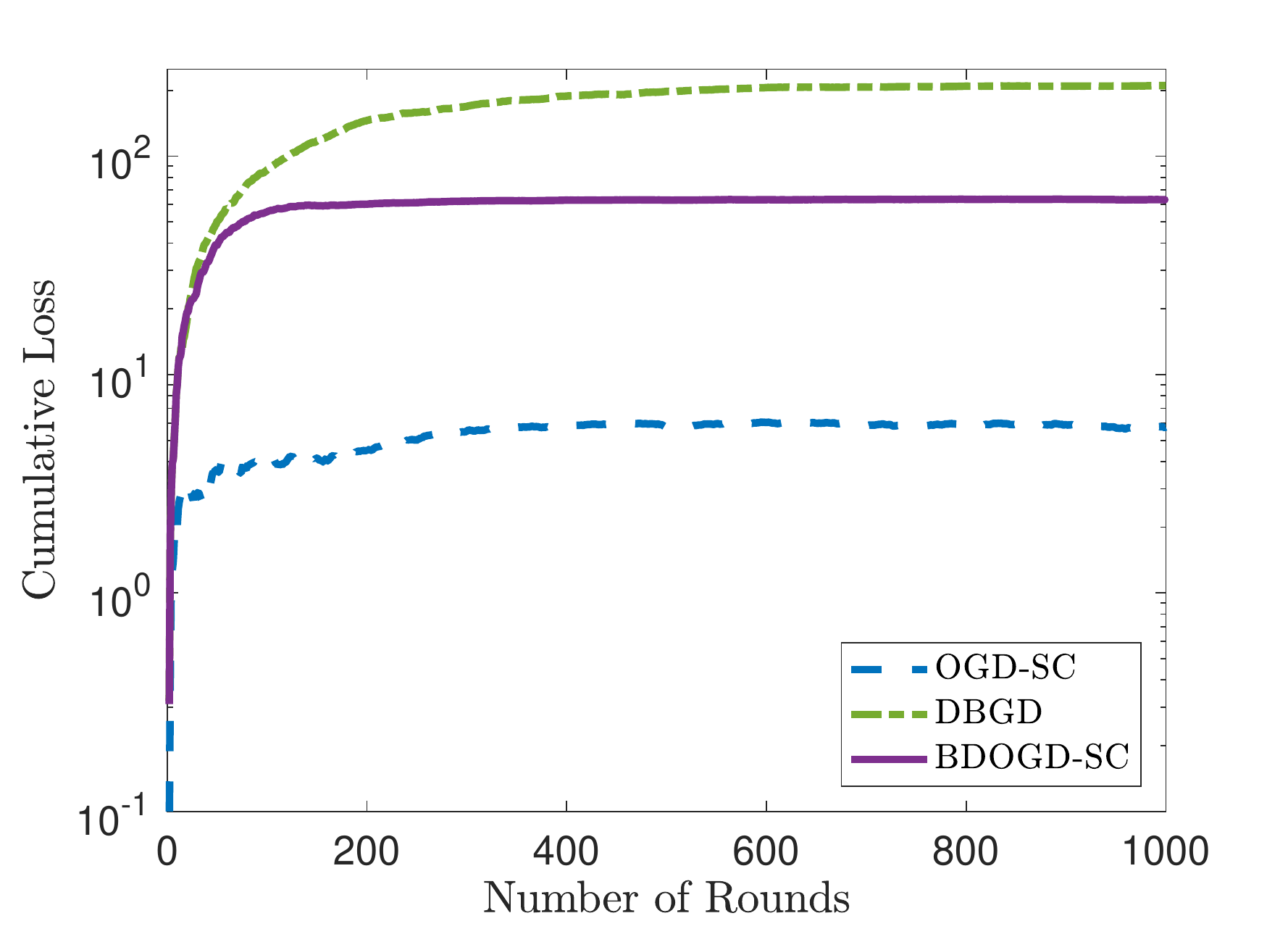}}
\caption{Comparisons of our Algorithm 2 against OGD-SC and DBGD.}
\label{fig2}
\end{figure*}

The experimental setup is inspired by \citet{Li_AISTATS19}. In each round $t$, the player chooses a decision $\x_t$ from the unit ball $\X=\{\x\in\mathbb{R}^{10}|\|\x\|\leq1\}$. Then, the loss function is generated as $f_t(\x)=\|\x\|^2+\mathbf{b}_t^\top\x$, where each element of $\mathbf{b}_t$ is uniformly sampled from $[-1,1]$. In this problem, the decision set $\X$ satisfies Assumption \ref{assum2} with $R=1$ and Assumption \ref{assumb1} with $r=1$. Each function $f_t(\x)$ is $2$-strongly convex and $2$-smooth, which satisfies Assumptions \ref{assum4} and \ref{assumb2}, respectively. Moreover, since $\nabla f_t(\x) = 2\x+\mathbf{b}_t$, we have
\[\|\nabla f_t(\x)\|\leq2\|\x\|+\|\mathbf{b}_t\|\leq 2+\sqrt{10}\]
for any $\x\in\X$, which implies that each function $f_t(\x)$ satisfies Assumption \ref{assum1} with $L=2+\sqrt{10}$.

We set $T=1000$, and consider two cases: the low delayed setting, in which the delays are periodically generated with length $2,3,2,1,4,1,3$, and the high delayed setting, in which the delays are periodically generated with length $20,30,20,10,40,10,30$. In the low delayed setting, the maximum delay is $d=4=O(1)$. In the other setting, the maximum delay $d=40$ is on the order of $O(\sqrt{T})$.

We compared DOGD-SC and BDOGD-SC against online gradient descent for strongly convex functions (OGD-SC) \citep{Hazan_2007}, DOGD \citep{Quanrud15} and DBGD \citep{Li_AISTATS19}. Specifically, OGD-SC is implemented without delay, and other algorithms are implemented with delayed feedback. The parameters of these algorithms are set as what their theoretical results suggest. For OGD-SC, we set the learning rate as $\eta_t=1/(\beta t)$, where $\beta=2$ in our experiments. For DOGD and DBGD, a constant learning rate $\eta=1/(L\sqrt{T+D})$ is used. Moreover, we set $\delta=\ln T/T$ for BDOGD-SC and set $\delta=1/(T+D)$ for DBGD. Furthermore, we initialize the decision as $\x_1=\mathbf{1}/\sqrt{10}$ for algorithms in the full information setting, and $\x_1=(1-\delta)\mathbf{1}/\sqrt{10}$ for algorithms in the bandit setting, where $\mathbf{1}$ denotes the vector with each entry equal 1.

Fig.~\ref{fig1} shows the cumulative loss for OGD-SC, DOGD and our DOGD-SC. We find that in both low and high delayed settings, our DOGD-SC is better than DOGD. Moreover, in the low delayed setting, the performance of our DOGD-SC is significantly better than DOGD, and close to OGD-SC. These results confirm that our DOGD-SC can utilize the strong convexity to achieve better regret. In our experiments, we also find that the performance of our BDOGD-SC is almost the same as that of DOGD-SC, and better than DBGD which is very close to DOGD. To make a clear presentation, we put the results of BDOGD-SC and DBGD in Fig.~\ref{fig2}, which confirms the theoretical guarantee of BDOGD-SC. Since BDOGD-SC and DBGD query the function at $11$ points per round, the average results are reported.

\section{Conclusion}
In this paper, we consider the problem of OCO with unknown delays, and present a variant of DOGD for strongly convex functions called DOGD-SC. According to our analysis, it enjoys a better regret bound of $O(d\log T)$ for strongly convex functions. Furthermore, we propose a bandit variant of DOGD-SC to handle the bandit setting, and achieve the same regret bound. Experimental results verify the performance of DOGD-SC and its bandit variant for strongly convex functions.





\vskip 0.2in
\bibliography{ref}

\newpage
\appendix

\section{Proof of Theorem \ref{thm2}}
This proof is inspired by the work of \citet{Agarwal2010_COLT}, which combined the $(n+1)$-point gradient estimator with OGD, and proved the average regret bound in the non-delayed setting. In this paper, we combine the $(n+1)$-point gradient estimator with DOGD-SC, and prove the average regret bound in the general delayed setting.

According to Assumption \ref{assum1}, for any $i=1,\cdots,n$, we have
\begin{align*}
f_t(\x_{t}+\delta\e_{i})\leq f_t(\x_t)+L\|\delta\e_{i}\|\leq f_t(\x_t)+L\delta
\end{align*}
which implies that
\begin{equation}
\label{thm2_eq0}
\begin{split}
&\frac{1}{n+1}\sum_{t=1}^T\sum_{i=0}^{n}f_t(\x_{t}+\delta\e_i)-\sum_{t=1}^Tf_t(\x)\\
\leq&\sum_{t=1}^Tf_t(\x_{t})+\sum_{t=1}^T\frac{nL\delta}{n+1}-\sum_{t=1}^Tf_t(\x)\\
\leq&\sum_{t=1}^Tf_t(\x_{t})-\sum_{t=1}^T(f_t((1-\delta/r)\x)-L\delta\|\x\|/r)+TL\delta\\
\leq&\sum_{t=1}^Tf_t(\x_{t})-\sum_{t=1}^Tf_t((1-\delta/r)\x)+\frac{TLR\delta}{r}+TL\delta.
\end{split}
\end{equation}
for any $\x\in\X$.

Then, we only need to upper bound $\sum_{t=1}^Tf_t(\x_{t})-\sum_{t=1}^Tf_t((1-\delta/r)\x)$. To this end, we start by defining \[\ell_t(\x)=f_t(\x)+(\tilde{\g}_t-\nabla f_t(\x_t))^\top \x.\]
It is easy to verify that $\ell_t(\x)$ is also $\beta$-strongly convex, and $\nabla \ell_t(\x_t)=\tilde{\g}_t$. Therefore, Algorithm \ref{alg2} is actually performing Algorithm \ref{alg1} on the functions $\ell_t(\x)$ over the decision set $\X_\delta$.

Moreover, under Assumptions \ref{assum1} and \ref{assumb2}, Lemma \ref{lemb111} shows
\begin{equation*}
\begin{split}
\|\tilde{\g}_t\|\leq \sqrt{n}L \text{ and }\|\tilde{\g}_t-\nabla f_t(\x_t)\|\leq\frac{\sqrt{n}\alpha\delta}{2}
\end{split}
\end{equation*}
which implies that
\[\|\nabla \ell_t(\x)\|\leq\|\nabla f_t(\x)\|+\|\tilde{\g}_t-\nabla f_t(\x_t)\|\leq L+\frac{\sqrt{n}\alpha\delta}{2}.\]
Define $\tilde{L}=L+\frac{\sqrt{n}\alpha\delta}{2}$. Applying Theorem \ref{thm1} to the functions $\ell_t(\x)$, for any $\x\in\X$, we have
\begin{equation*}
\begin{split}
&\sum_{t=1}^T\ell_t(\x_{t})-\sum_{t=1}^T\ell_t((1-\delta/r)\x)\\
\leq&\sum_{t=1}^T\ell_t(\x_{t})-\min_{\x^\prime\in\X_\delta}\sum_{t=1}^T\ell_t(\x^\prime)\\
\leq&\left(4\beta R\tilde{L}+5\tilde{L}^2\right)\frac{d}{\beta}\left(1+\ln\frac{T}{|\F_s|}\right).
\end{split}
\end{equation*}
Furthermore, for any $\x\in\X$, we have
\begin{equation*}
\begin{split}
&\sum_{t=1}^Tf_t(\x_{t})-\sum_{t=1}^Tf_t((1-\delta/r)\x)\\
=&\sum_{t=1}^T\ell_t(\x_{t})-\sum_{t=1}^T\ell_t((1-\delta/r)\x)+\sum_{t=1}^T(\tilde{\g}_t-\nabla f_t(\x_t))^\top (\x_t-(1-\delta/r)\x)\\
\leq&\sum_{t=1}^T\ell_t(\x_{t})-\sum_{t=1}^T\ell_t((1-\delta/r)\x)+\sum_{t=1}^T\|\tilde{\g}_t-\nabla f_t(\x_t)\|\|\x_t-(1-\delta/r)\x\|\\
\leq&\left(4\beta R\tilde{L}+5\tilde{L}^2\right)\frac{d}{\beta}\left(1+\ln\frac{T}{|\F_s|}\right)+\sum_{t=1}^T\sqrt{n}\alpha\delta R.
\end{split}
\end{equation*}
Combining with (\ref{thm2_eq0}), for any $\x\in\X$, we have
\begin{align*}
&\frac{1}{n+1}\sum_{t=1}^T\sum_{i=0}^{n}f_t(\x_{t}+\delta\e_i)-\sum_{t=1}^Tf_t(\x)\\
\leq&\left(4\beta R\tilde{L}+5\tilde{L}^2\right)\frac{d}{\beta}\left(1+\ln\frac{T}{|\F_s|}\right)+\sum_{t=1}^T\sqrt{n}\alpha\delta R+\frac{TLR\delta}{r}+TL\delta.
\end{align*}
We complete this proof by substituting $\delta=\frac{c\ln T}{T}$ into the above inequality.
\begin{algorithm}[t]
\caption{A Bandit Variant of DOGD-SC with Two Queries per Round}
\label{alg3}
\begin{algorithmic}[1]
\STATE \textbf{Input:} A parameter $\delta> 0$
\STATE \textbf{Initialization:} Choose an arbitrary vector $\x_1\in\X_\delta$ and set $h_0=0$
\FOR{$t=1,2,\cdots,T$}
\STATE Sample $\u_t\sim\SS^n$
\STATE Query $f_t(\x_{t}+\delta\u_t),f_t(\x_{t}-\delta\u_t)$
\STATE $h_t=h_{t-1}+\frac{|\F_t|\beta}{2}$
\STATE $\x_{t+1}=\left\{
\begin{aligned}
&\Pi_{\X_\delta}\left(\x_{t}-\frac{1}{h_t}\sum_{k\in\F_t}\tilde{\g}_k\right)\text{ if } |\F_t|>0\\
&\x_{t}\quad\quad\quad\quad\quad\quad\quad\quad\quad\text{ otherwise}
\end{aligned}\right.$
 where $\tilde{\g}_k=\frac{n}{2\delta}(f_k(\x_{k}+\delta\u_k)-f_k(\x_k-\delta\u_k))\u_k$
\ENDFOR
\end{algorithmic}
\end{algorithm}
\section{A Bandit Variant of DOGD-SC with Two Queries per Round}
In Section 3.2, we have proposed BDOGD-SC for the bandit setting, which requires $n+1$ queries per round. To reduce the number of queries, we further present a bandit variant of DOGD-SC with only two queries per round, for the case where the time stamp of each delayed feedback is known.

According to \citet{Agarwal2010_COLT}, for a function $f(\x):\X\mapsto\mathbb{R}$ and a point $\x\in\X_\delta$, the two-point gradient estimator queries \[f(\x+\delta\u),f(\x-\delta\u)\] where $\u$ is uniformly at random sampled from the unit sphere $\SS^n$, and estimates the gradient $\nabla f(\x)$ by
\begin{equation}
\label{eq_est_grad2}
\tilde{\g}=\frac{n}{2\delta}(f(\x+\delta\u)-f(\x-\delta\u))\u.
\end{equation}
Combining DOGD-SC with this technique, a new bandit variant of DOGD-SC is outlined in Algorithm \ref{alg3}, and named as a bandit variant of DOGD-SC with two queries per round. Specifically, given $0<\delta<r$ and $\X_\delta=(1-\delta/r)\X$, in each round $t\in[T]$, the learner queries $f_t(\x_{t}+\delta\u_t),f_t(\x_{t}-\delta\u_t)$, where $\x_t\in \X_\delta$ and $\u_t$ is uniformly at random sampled from the unit sphere $\SS^n$. 
After receiving the feedback
\[\left\{f_k(\x_{k}+\delta\u_k),f_k(\x_{k}-\delta\u_k)|k+d_k-1=t\right\}\]
we can compute the approximate gradient \[\tilde{\g}_k=\frac{n}{2\delta}(f_k(\x_{k}+\delta\u_k)-f_k(\x_k-\delta\u_k))\u_k.\]
for any $k\in\F_t$ according to (\ref{eq_est_grad2}), which needs to use the time stamp $k$ to match the feedback $\{f_k(\x_{k}+\delta\u_k),f_k(\x_{k}-\delta\u_k)\}$ with the random vector $\u_k$. Then, we compute the sum $\sum_{k\in\F_t}\tilde{\g}_k$, and update $\x_t$ as
\[\x_{t+1}=\left\{
\begin{aligned}
&\Pi_{\X_\delta}\left(\x_{t}-\frac{1}{h_t}\sum_{k\in\F_t}\tilde{\g}_k\right)\text{ if } |\F_t|>0,\\
&\x_{t}\quad\quad\quad\quad\quad\quad\quad\quad\quad\text{ otherwise.}
\end{aligned}\right.\]
Following previous studies for the bandit setting \citep{OBO05,Saha_2011}, we assume that the adversary is oblivious, and establish the following theorem.
\begin{thm}
\label{thm3}
Let $\x$ be an arbitrary vector in the set $\X$. Let $\x_{t,1}=\x_t+\delta\u_t$ and $\x_{t,2}=\x_t-\delta\u_t$. Define $\tilde{L}=2L+Ln$. Let $\delta=\frac{c\ln T}{T}$, where $c>0$ is a constant such that $\delta<r$. Under Assumptions \ref{assum1}, \ref{assum2}, \ref{assum4} and \ref{assumb1}, Algorithm \ref{alg2} ensures
\begin{align*}
\E\left[\frac{1}{2}\sum_{t=1}^T\sum_{i=1}^2f_t(\x_{t,i})-\sum_{t=1}^Tf_t(\x)\right]\leq&\left(4\beta R\tilde{L}+5\tilde{L}^2\right)\frac{d}{\beta}\left(1+\ln\frac{T}{|\F_s|}\right)+3cL\ln T+\frac{RL\ln T}{r}.
\end{align*}
where $s=\min\left\{t|t\in[T+d-1],|\F_t|>0\right\}$.
\end{thm}

\section{Proof of Theorem \ref{thm3}}
This proof is inspired by the work of \citet{Agarwal2010_COLT}, which analyzed the expected regret for the combination of the two-point gradient estimator and OGD in the non-delayed setting.

We first introduce the $\delta$-smoothed version of a function $f(\x)$ and the corresponding properties, which will be used in the following proof. For a function $f(\x)$, its $\delta$-smoothed version is defined as \[\hat{f}(\x)=\mathbb{E}_{\u\sim\B^n}[f(\x+\delta\u)]\]
and satisfies the following two lemmas.
\begin{lem}
\label{smoothed_lem2}
(Lemma 1 in \citet{OBO05})
Let $\delta>0$, we have
\[\nabla\hat{f}(\x)=\mathbb{E}_{\u\sim\SS^n}\left[\frac{n}{\delta}f(\x+\delta\u)\u\right]\]
where $\SS^n$ denotes the unit sphere in $\mathbb{R}^n$.
\end{lem}
\begin{lem}
\label{smoothed_lem1}
(Derived from Lemma 2.6 of \citet{Hazan2016})
Let $f(\x):\mathbb{R}^n\to\mathbb{R}$ be $\beta$-strongly convex and $L$-Lipschitz over a convex and compact set $\X\subset\mathbb{R}^n$. Then, $\hat{f}(\x)$ has the following properties.
\begin{itemize}
\item $\hat{f}(\x)$ is $\beta$-strongly convex over $\X_\delta$;
\item $|\hat{f}(\x)-f(\x)|\leq\delta L$ for any $\x\in\X_\delta$;
\item $\hat{f}(\x)$ is $L$-Lipschitz over $\X_\delta$.
\end{itemize}
\end{lem}
Let $\hat{\x}=(1-\delta/r)\x$. We have
\begin{equation}
\label{thm3_eq1}
\begin{split}
&\frac{1}{2}\sum_{t=1}^T\sum_{i=1}^2f_t(\x_{t,i})-\sum_{t=1}^Tf_t(\x)\\
=&\frac{1}{2}\sum_{t=1}^T(f_t(\x_t+\delta\u_t)+f_t(\x_t-\delta\u_t))-\sum_{t=1}^Tf_t(\x)\\
\leq&\frac{1}{2}\sum_{t=1}^T(f_t(\x_t)+L\|\delta\u_t\|+f_t(\x_t)+L\|\delta\u_t\|)-\sum_{t=1}^T(f_t(\hat{\x})-L\delta\|\x\|/r)\\
\leq&\sum_{t=1}^Tf_t(\x_t)-\sum_{t=1}^Tf_t(\hat{\x})+LT\delta+\frac{RLT\delta}{r}\\
\leq&\sum_{t=1}^T(\hat{f}_t(\x_t)+\delta L)-\sum_{t=1}^T(\hat{f}_t(\hat{\x})-\delta L)+LT\delta+\frac{RLT\delta}{r}\\
=&\sum_{t=1}^T\hat{f}_t(\x_t)-\sum_{t=1}^T\hat{f}_t(\hat{\x})+3LT\delta+\frac{RLT\delta}{r}
\end{split}
\end{equation}
where the first inequality is due to Assumption \ref{assum1} and the last inequality is due to Lemma \ref{smoothed_lem1}.

Then, we only need to upper bound $\sum_{t=1}^T\hat{f}_t(\x_t)-\sum_{t=1}^T\hat{f}_t(\hat{\x})$. Similar to the proof of Theorem \ref{thm2}, we define
\[\ell_t(\x)=\hat{f}_t(\x)+(\tilde{\g}_t-\nabla \hat{f}_t(\x_t))^\top \x.\]
According to Lemma \ref{smoothed_lem2}, we have
\begin{align*}
\E_{\u_t}[\tilde{\g}_t]=&\E_{\u_t}\left[\frac{n}{2\delta}(f_t(\x_{t}+\delta\u_t)-f_t(\x_t-\delta\u_t))\u_t\right]\\
=&\E_{\u_t}\left[\frac{n}{\delta}f_t(\x_{t}+\delta\u_t)\u_t\right]=\nabla\hat{f}_t(\x_t)
\end{align*}
where the second equality is due to the fact that the distribution of $\u_t$ is symmetric.

Then, we have $\E_{\u_t}[\tilde{\g}_t-\nabla \hat{f}_t(\x_t)]=0$, which implies that
\begin{equation}
\label{thm3_eq2}
\E\left[\sum_{t=1}^T(\hat{f}_t(\x_t)-\hat{f}_t(\hat{\x}))\right]=\E\left[ \sum_{t=1}^T(\ell_t(\x_t)-\ell_t(\hat{\x}))\right].
\end{equation}
Therefore, we only need to derive an upper bound of $\sum_{t=1}^T\ell_t(\x_t)-\sum_{t=1}^T\ell_t(\hat{\x})$.

According to the definition of $\ell_t(\x)$, it is easy to verify that $\nabla\ell_t(\x_t)=\tilde{\g}_t$. Moreover, from Lemma \ref{smoothed_lem1}, $\hat{f}_t(\x)$ is $\beta$-strongly convex, which implies that $\ell_t(\x)$ is also $\beta$-strongly convex. Therefore, Algorithm \ref{alg3} is actually performing Algorithm \ref{alg1} on the functions $\ell_t(\x)$ over the decision set $\X_\delta$.

Before using Theorem \ref{thm1}, we need to prove that $\ell_t(\x)$ is also Lipschitz. From Lemma \ref{smoothed_lem1}, $\hat{f}_t(\x)$ is $L$-Lipschitz. So, for any $\x,\y\in\X_\delta$, it is not hard to verify that
\begin{align*}
|\ell_t(\x)-\ell_t(\y)|\leq&|\hat{f}_t(\x)-\hat{f}_t(\y)|+|(\tilde{\g}_t-\nabla \hat{f}_t(\x_t))^\top (\x-\y)|\\
\leq&L\|\x-\y\|+\|\tilde{\g}_t-\nabla \hat{f}_t(\x_t)\|\|\x-\y\|\\
\leq&(L+\|\tilde{\g}_t\|+\|\nabla \hat{f}_t(\x_t)\|)\|\x-\y\|\\
\leq&(2L+Ln)\|\x-\y\|
\end{align*}
where the last inequality is due to $\|\nabla \hat{f}_t(\x_t)\|\leq L$ and
\begin{align*}
\|\tilde{\g}_t\|=&\frac{n}{2\delta}\|(f_t(\x_{t}+\delta\u_t)-f_t(\x_t-\delta\u_t))\u_t\|\\
=&\frac{n}{2\delta}|f_t(\x_{t}+\delta\u_t)-f_t(\x_t-\delta\u_t)|\\
\leq&\frac{n}{2\delta}L\|2\delta\u_t\|=nL.
\end{align*}
Let $\tilde{L}=2L+Ln$. Since $\ell_t(\x)$ is $\beta$-strongly convex and $\tilde{L}$-Lipschitz. Applying Theorem \ref{thm1} to the functions $\ell_t(\x)$, we have
\begin{equation}
\label{thm3_eq3}
\begin{split}
\sum_{t=1}^T(\ell_t(\x_t)-\ell_t(\hat{\x}))\leq\left(4\beta R\tilde{L}+5\tilde{L}^2\right)\frac{d}{\beta}\left(1+\ln\frac{T}{|\F_s|}\right).
\end{split}
\end{equation}
Combining (\ref{thm3_eq1}), (\ref{thm3_eq2}) and (\ref{thm3_eq3}), we have
\begin{align*}
&\E\left[\frac{1}{2}\sum_{t=1}^T\sum_{i=1}^2f_t(\x_{t,i})-\sum_{t=1}^Tf_t(\x)\right]\\
\leq&\E\left[\sum_{t=1}^T(\hat{f}_t(\x_t)-\hat{f}_t(\hat{\x}))\right]+3LT\delta+\frac{RLT\delta}{r}\\
=&\E\left[ \sum_{t=1}^T(\ell_t(\x_t)-\ell_t(\hat{\x}))\right]+3LT\delta+\frac{RLT\delta}{r}\\
\leq&\left(4\beta R\tilde{L}+5\tilde{L}^2\right)\frac{d}{\beta}\left(1+\ln\frac{T}{|\F_s|}\right)+3LT\delta+\frac{RLT\delta}{r}.
\end{align*}
We complete this proof by substituting $\delta=\frac{c\ln T}{T}$ into the above inequality.

\section{Proof of Lemma \ref{smoothed_lem1}}
The first and second properties have been presented in Lemma 2.6 of \citet{Hazan2016}. The last property is proved by
\begin{align*}
|\hat{f}(\x)-\hat{f}(\y)|=&\left|\mathbb{E}_{\u\sim\B^d}[f(\x+\delta\u)]-\mathbb{E}_{\u\sim\B^d}[f(\y+\delta\u)]\right|\\
=&\left|\mathbb{E}_{\u\sim\B^d}[f(\x+\delta\u)-f(\y+\delta\u)]\right|\\
\leq&\mathbb{E}_{\u\sim\B^d}[|f(\x+\delta\u)-f(\y+\delta\u)|]\\
\leq&\mathbb{E}_{\u\sim\B^d}[L\|\x-\y\|]\\
=&L\|\x-\y\|
\end{align*}
where the first inequality is due to Jensen's inequality, and the second inequality is due to the fact that $f(\x)$ is $L$-Lipschitz over $\X$.

\end{document}